\documentclass{article}
\usepackage{spconf,amsmath,graphicx}
\usepackage[usenames, dvipsnames]{color}
\usepackage{hyperref}

\title{ICON: An Interactive Approach to Train Deep Neural Networks for Segmentation of Neuronal Structures}
\name{F. Gonda$^{\star}$, V. Kaynig$^{\star}$, R. Thouis$^{\star}$, D. Haehn$^{\star}$, J. W. Lichtman$^{\dagger}$, T. Parag$^{\star}$, H. Pfister$^{\star}$\thanks{}}
\address{$^{\star}$ Harvard University, John A. Paulson School of Engineering and Applied Sciences,\\ Cambridge, Massachusetts, USA.
$^{\dagger}$ Harvard University, Department of Molecular and Cellular\\ 
             Biology and Center for Brain Science, Cambridge, Massachusetts, USA.\\
             ${\lbrack}$fgonda, vkaynig, thouis, haehn, paragt, pfister${\rbrack}$@g.harvard.edu, jeff@mcb.harvard.edu
}

\begin{document}
%
\maketitle
\begin{abstract}
We present an interactive approach to train a deep neural network pixel classifier for the segmentation of neuronal structures. An interactive training scheme reduces the extremely tedious manual annotation task that is typically required for deep networks to perform well on image segmentation problems.  Our proposed method employs a feedback loop that captures sparse annotations using a graphical user interface, trains a deep neural network based on recent and past annotations, and displays the prediction output to users in almost real-time. Our implementation of the algorithm also allows multiple users to provide annotations in parallel and receive feedback from the same classifier. Quick feedback on classifier performance in an interactive setting enables users to identify and label examples that are more important than others for segmentation purposes.  Our experiments show that an interactively-trained pixel classifier produces better region segmentation results on Electron Microscopy (EM) images than those generated by a network of the same architecture trained offline on exhaustive ground-truth labels.
\end{abstract}
\section{Introduction}
\label{sec:intro}
Connectomics is an emerging discipline of neuroscience dedicated to the reconstruction of neural structures and connectivity from brain images. Electron Microscopy (EM) can provide extremely detailed images (at nanomemter scale) of animal brain tissues for a comprehensive neural reconstruction. Recording at such high resolution generates a massive amount of data from a relatively small brain region; a ${{1mm}^{3}}$ volume of rat cortex amounts to 33,333 images of size $250,000 \times 25,000$ when imaged at $4 \times 4 \times 30$ nm $x, y, z$ resolution. Automated or semi-automated processing is the most viable strategy to process datasets at this scale.

Segmentation of neuron regions in EM volumes has become the central tool for many (semi-) automated neural reconstruction efforts \cite{Helmstaedter, Takemura03112015, Kaynig2015}. Almost all these segmentation approaches utilize a pixel classifier to distinguish cell boundary (or membrane) pixels from cell interior (or other organelle) pixels. A set of annotated pixels is required to train such a classifier. Most of the existing segmentation algorithms \cite{jain10cvpr, Ciresan2011} require an exhaustive ground-truth volume, where each pixel (or voxel) within the volume is annotated by expert users for training. The type of annotation is generally a label for boundary pixels.

Exhaustive annotation of a sufficiently large volume demands significant time and effort from an expert in cell biology.  Our experience suggests a ${1024}$ x ${1024}$ x ${250}$ EM volume would require $6 \sim 8$ weeks of dedicated labeling effort from a neurobiologist. Such a manual labeling step becomes a substantial bottleneck for the overall neural reconstruction process. This impediment is compounded for large datasets collected from multiple brain areas with biologically different cell characteristics and difference in tissue preparation techniques. 

Several researchers in the EM segmentation community realized this issue and proposed different algorithms for training pixel classifiers from a relatively small (and often sparse) set of training examples. The interactive software Ilastik \cite{Sommer2011} for learning a Random Forest pixel detector has become very popular in the bioinformatics community. Rather than asking for dense pixelwise labels, Ilastik provides a user interface to identify training examples that could potentially improve the classifier performance given the current classifier output overlaid on the input image. The works of Kaynig et. al. \cite{Kaynig2015} and Parag et.al. \cite{Parag2015} also proposed methods for sparse selection of a subset of training examples and demonstrated their advantages for EM segmentation. However, the strong dependence of the random forest classifiers on hand-tuned features has the potential to limit their performances on images from different EM preparation/imaging techniques and, more generally, from other data modalities.

In this paper, we present a method for interactive training of a Convolutional Neural Network (CNN) classifier \cite{Ciresan2011} for EM image segmentation.  The user paints on an input image to mark the pixels corresponding to a particular class. Presented with the pixel classification performance of the CNN, the user marks one or more areas that s/he thinks would improve the quality of the segmentation. We propose an efficient training algorithm to yield near real-time feedback to the user. Our training method (Section \ref{ssec:training}) is designed to emphasize the most recent user input while retraining the CNN. In addition, the learning algorithm also keeps track of past training examples that were adversely affected by this update and prioritizes them for learning in the next iteration. For segmentation purposes, our training strategy enables us to train a CNN with less than $2\%$ of the total training examples, that is a few hundred thousands out of millions of total training examples.  Compared to a CNN of the same architecture trained on all ground-truth images, our technique achieves slightly better segmentation accuracy. Our results corroborate well with past studies \cite{Kaynig2015, Parag2015} that found that interactive training tends to produce a classifier that performs better for segmentation. 

The primary objective of training the pixel detector using our method is region segmentation. Therefore, we use Variation of Information (VI), which has become a standard measure in connectomics~\cite{Kaynig2015, Nunez-Iglesias13,Parag2015}, to quantitatively compare segmentation performance.  To our knowledge, ours is the first effort for interactive training of deep networks for EM segmentation. Although the method is primarily targeted and tested on EM data, we believe several problems in biomedical image segmentation could benefit from our method, including mitosis detection~\cite{ciresan2013miccai}, cell segmentation on histopathology images~\cite{su2015miccai}, and cell tracking \cite{maska14bioinformatics}. 

\section{Method}
\label{sec:methods}
\begin{figure}[htb]
\begin{minipage}[b]{1.0\linewidth}
  \centering  \centerline{\includegraphics[clip, trim=4.0cm 2.5cm 4.0cm 3.8cm, width=8.5cm]{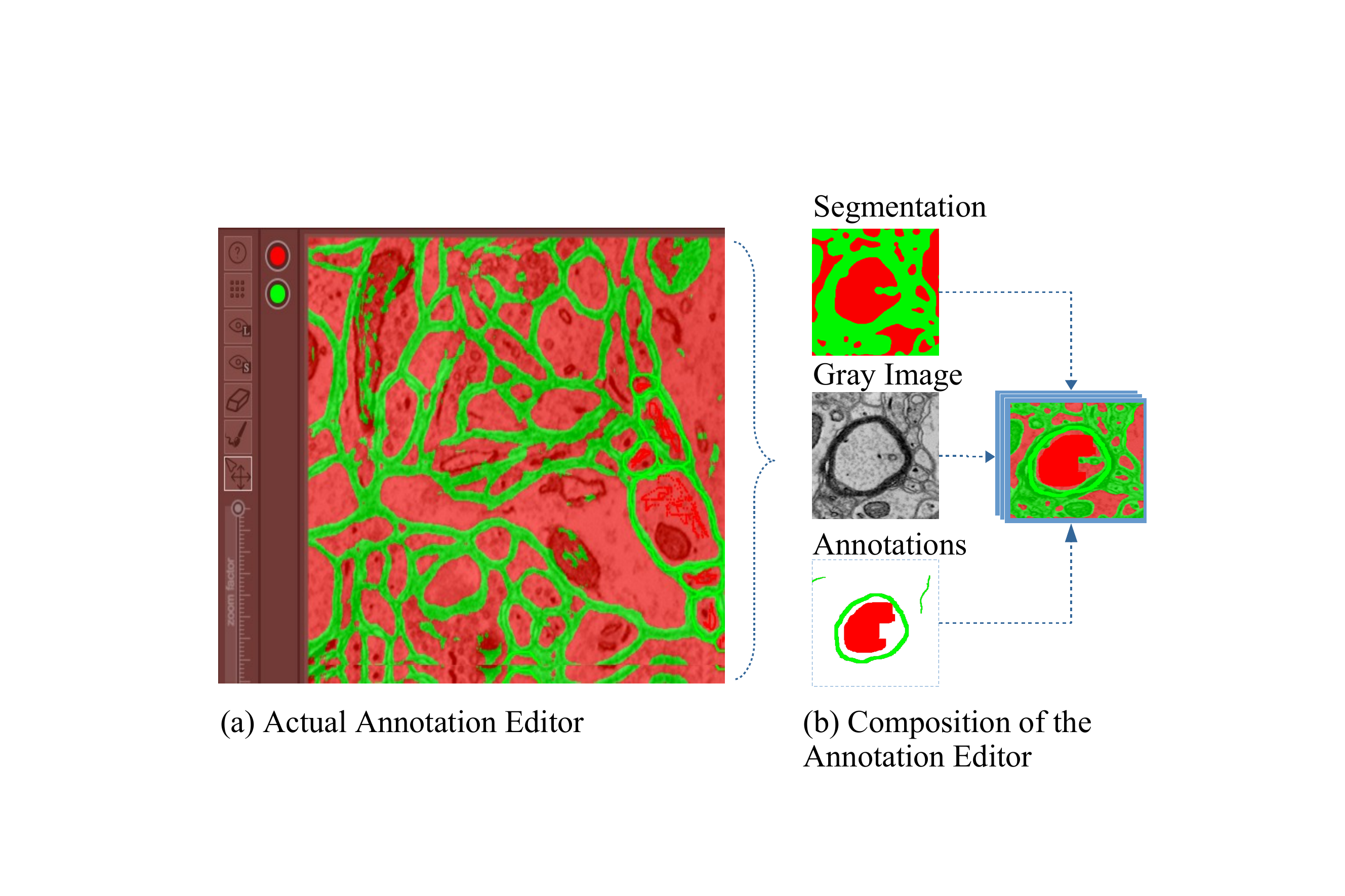}}
\end{minipage}
\caption{(a) ICON's main screen for capturing annotation and visualizing classifier output.  It consists of tools for editing annotations, controlling visualizations, and radio buttons to indicate object classes. (b) The three layers that make up the visualization portion of the screen.}
\label{fig:iconui}
\end{figure}

ICON, our tool for boundary detection, employs a CNN as a pixel detector to classify each pixel into membrane and non-membrane classes in a cell. The CNN is trained with sparse annotations collected interactively from users over a web-based graphical user interface (GUI) shown in Fig. \ref{fig:iconui}. 

The annotations are saved in a central database as training samples to be exploited for training by the CNN.  The CNN confidences for pixel detection are overlaid on the grayscale input images and displayed on the GUI to guide the user during the annotation process. Based on the performance of the pixel detector and the user expertise, the user annotates locations where an improvement in pixel classification can lead to more accurate segmentation.
 
The CNN classifier runs on two parallel threads on a separate compute node than the GUI, as described in Section \ref{ssec:architecture}.  The first thread is dedicated to training. It draws samples from the central database to train the classifier and saves to disk a model when better validation accuracy is achieved.  The saved model is used by a second thread that produces segmentation outputs on demand for images that are currently being annotated.

The classifier is trained on samples from multiple images collected from users via a web interface. Because we are training the classifier from scratch, a minimum of 100,000 samples are required before useful feedback is seen. Our training method is described next. 

\subsection{Classifier Training}
\label{ssec:training}
\begin{figure}[htb]
\begin{minipage}[b]{1.0\linewidth}
  \centering
  \centerline{\includegraphics[clip, trim=4.5cm 6.5cm 4.5cm 8.6cm, width=8.5cm]{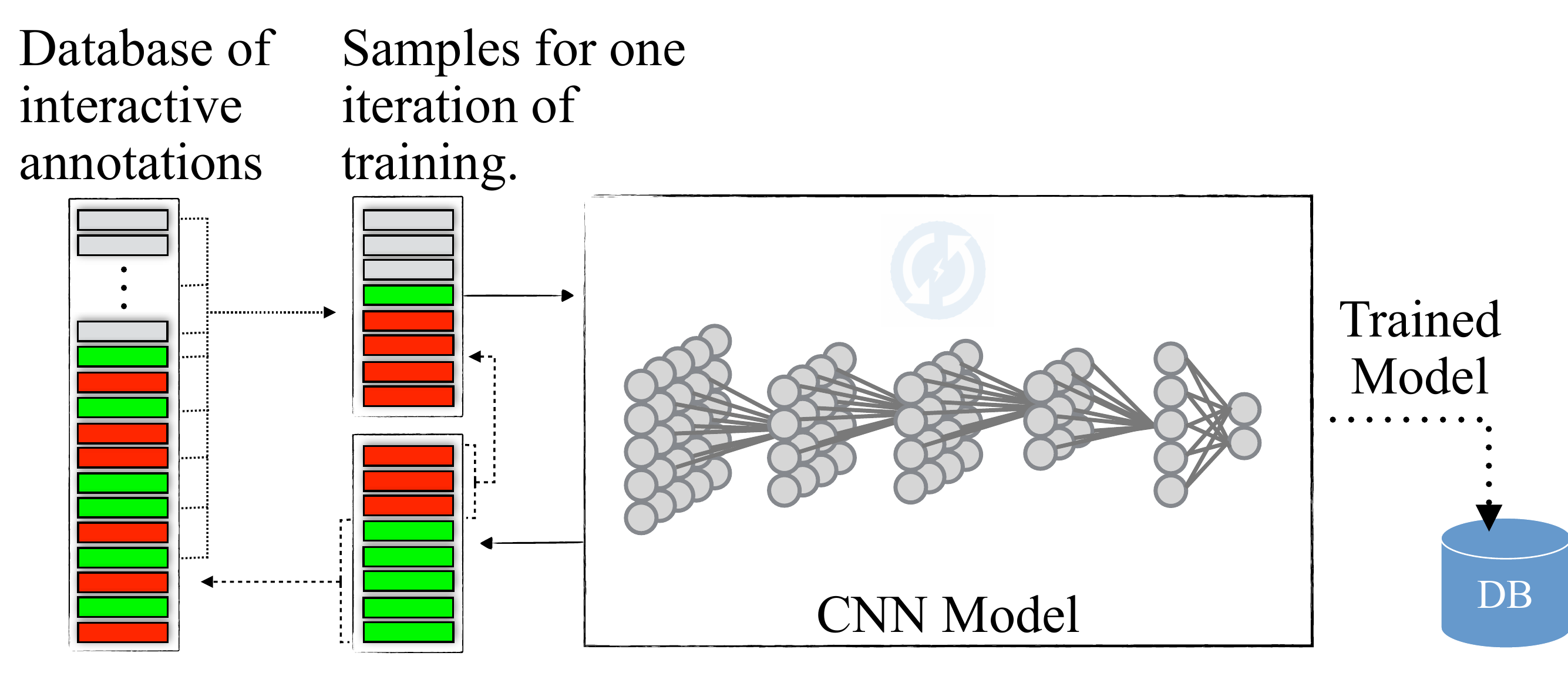}}
\end{minipage}
\caption{Iterative classifier training.  New samples are shown in gray.  Good performing samples are shown in green.  Poorly-performing samples are shown in red. }
\label{fig:training}
\end{figure}

One of the issues with conventional CNN training in an interactive environment is how to achieve real-time feedback to keep up with annotation input from users.  In our training strategy, depicted in Fig. \ref{fig:training}, we focus on accelerating the learning process so that the classifier is able to produce segmentation output in a reasonable amount of time. In addition, our learning scheme also employs a training sample selection technique in order to be responsive to the user input and maintain a level of accuracy over all the samples annotated so far.

At the beginning of a training iteration, the learning thread draws a large number of samples from the central database of annotations. A \textit{sample} is a patch of the image extracted from a square region centered around a pixel. The samples are drawn across all training images and labels, and we sub-sample classes equally to avoid imbalanced training data. During the sampling process, priority is given to new annotations. We also perform arbitrary rotation on the samples to ensure the network can recognize objects in different orientations.

After sampling, we combine the training samples with poorly-performing examples from the previous iteration of training.  The performance of a sample is measured by the the error between the network output and the target value.  A large error indicates the sample has not been learned by the network, therefore the system needs to present this sample more frequently to the network.  Given the set of all samples ${S}$, the set of poorly performing samples is described as: $S_b = \{x_i \in S ~|~ || y_i - f(x_i) || > \delta  \}$, where ${y_i}$ is the label of example ${x_i}$, ${f_i}$  refers to the network output, and ${\delta}$ is a threshold that is set by application to ${0.5}$ since the network is classifying membranes and non-membranes.  Next, the learning thread iterates over the samples and perform Stochastic Gradient Descent (SGD) training using a small subset of the samples at a time.  

After each training iteration, the samples are evaluated and a maximum of  50\% of the poorly-performing samples is retained for the next iteration of training.  

\subsection{Classifier Refinement}
\label{ssec:annotations}
\begin{figure}[htb]
\begin{minipage}[b]{1.0\linewidth}
  \centering
  \centerline{\includegraphics[clip, trim=4.5cm 7.5cm 5.5cm 7.5cm, width=8.5cm]{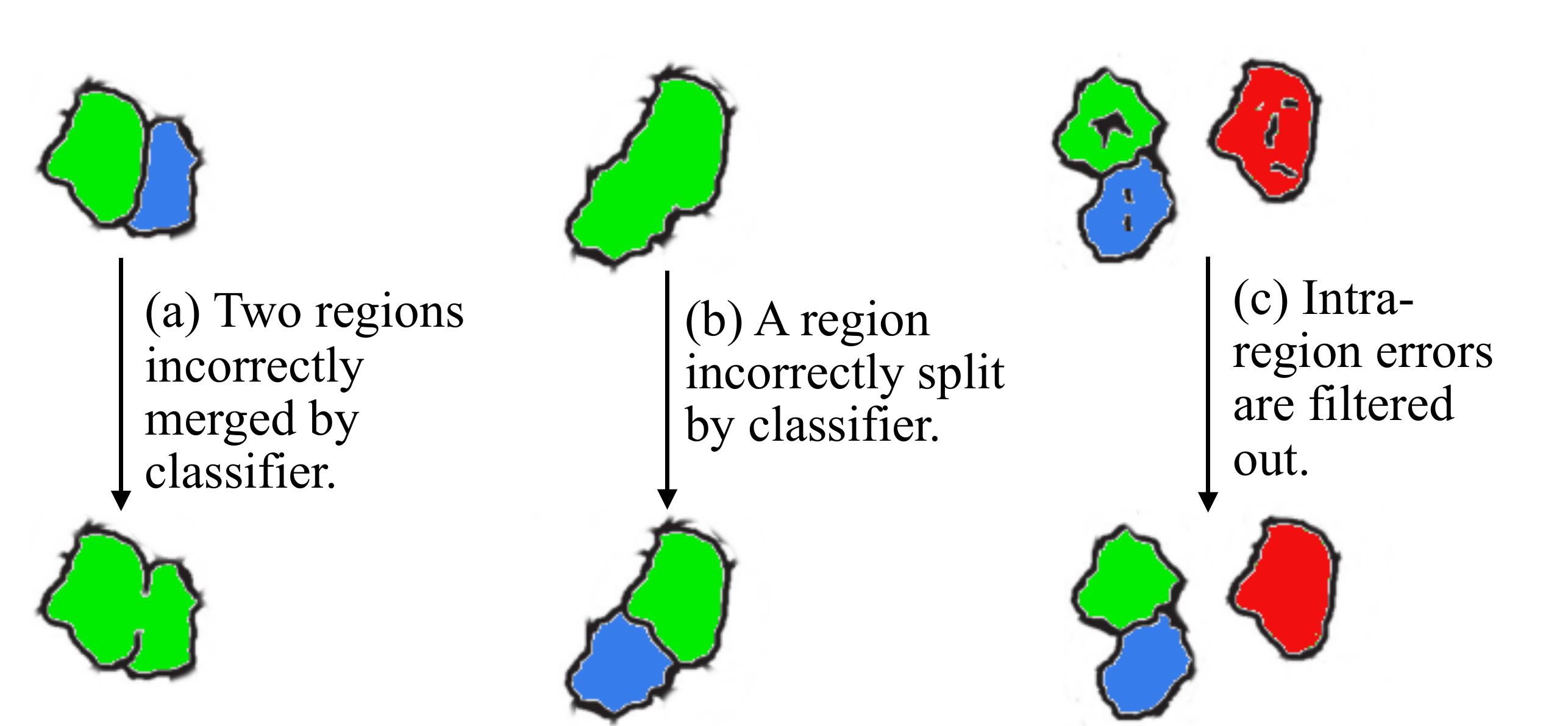}}
\end{minipage}
\caption{Segmentation error induced by pixel classification: (a) A merge error, where two regions are incorrectly combined into one. (b) A split error, where one region is incorrectly divided into two. (c) Pixel classification error that does not affect segmentation.}
\label{fig:classifiererrors}
\end{figure}

Among the classification errors generated by the deep network, it is important to identify the misclassified pixels that causes false merge and splits in the segmentation. We display examples of these scenarios respectively in Fig. \ref{fig:classifiererrors}(a) and (b). Misclassification of pixels within segmented regions -- as shown in Fig. \ref{fig:classifiererrors}(c) -- often do not affect the quality of segmentation as a result of the region growing methods used in subsequent processing steps of the pixel predictions \cite{Kaynig2015}.

Therefore, during the annotation process, we use the pixel confidences as an overlay to focus the manual effort on fixing classifier mistakes in order to guide the classifier to pay more attention to these cases.  This refinement process is important because we are interested in region segmentation and we evaluate our system on this basis.

\subsection{System Implementation}
\label{ssec:architecture}
\begin{figure}[htb]
\begin{minipage}[b]{1.0\linewidth}
  \centering
  \centerline{\includegraphics[clip, trim=6.5cm 7.0cm 7.8cm 6.2cm, width=8.5cm]{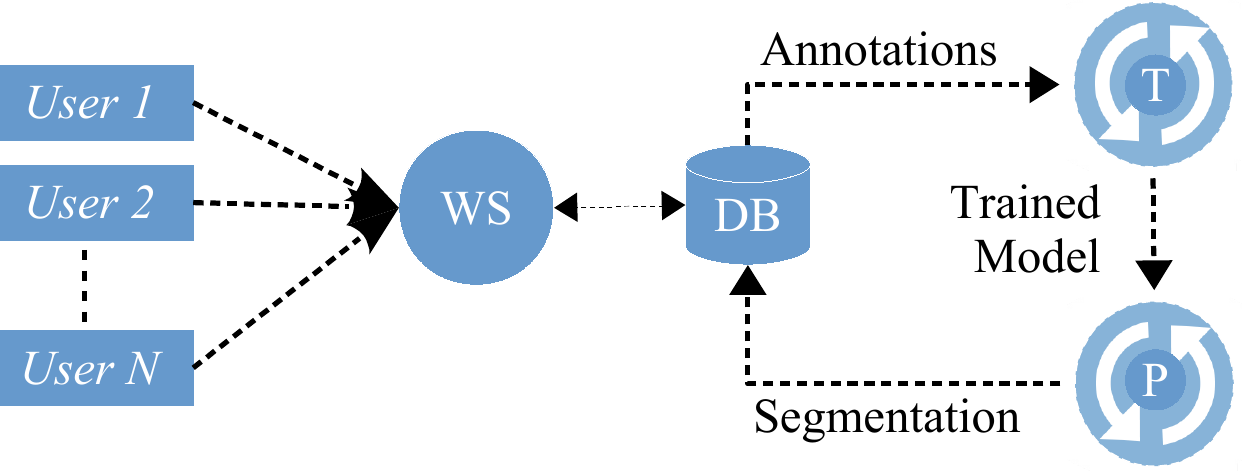}}
\end{minipage}
\caption{The architecture of the system consists of a GUI, a web service (WS), a database (DB), a training thread (T), and a prediction thread (P).}
\label{fig:architecture}
\end{figure}

The architecture of the system is comprised of several parts, depicted in Fig. \ref{fig:architecture}.  On the front-end is a GUI that runs in a web browser and provides facilities for editing annotations. The GUI is connected to a web service that runs on a server.  The web service facilitates the data exchange between the GUI and the CNN model by utilizing a database to store annotations and retrieve segmentation outputs.  The classifier is trained on one thread and outputs a trained model that is used by a prediction thread. The prediction thread runs in parallel to the training thread and produces probability maps as described in section \ref{ssec:training}.  These parts are integrated into a real-time feedback loop that synchronizes the server with user-provided annotations and the GUI with segmentation outputs from the classifier.

\section{Results}
\label{sec:results}
To evaluate our system, we use data from sections of a tissue of a dense mammalian neuropil from layers 4 and 5 of the S1 primary somatosensory cortex of a 5 month old healthy C45BL/6J mouse.  240 gray-value section images with $1024 \times 1024$ pixels are used for training and validation.  For testing, we use 120 images divided into two sets from which results were generated.  Each image has a corresponding ground-truth membrane and background labels for every pixel.  

The CNN architecture consists of two convolutional layers, each composed of 48 filters of size ${5 \times 5}$. The fully-connected layer of the network consists of 200 units and the output layer has two units, one for membrane and the other for non-membrane.  We use a learning rate of 0.01 and a momentum of 0.9. 

For comparison, we trained another CNN with the same architecture as the interactive setting  offline with the full set of ground-truth labels. The interactive classifier was trained on ${185,990}$ pixels sparsely annotated on ten training images, i.e.,\ 1.7\% of total pixels across the ten images. The interactive classifier typically converges after one hour of training time and the offline classifier after two hours. 

We use VI to quantitatively measure segmentation performances produced on the test images by the different classifiers.  We first generate probability maps for each image.  Then we threshold each probability map at different intervals. The threshold is a decision boundary value that is applied to the probability map to separate membranes from non-membranes. We then compute region clusters from the probability maps using Mahotas \cite{mahotas} connected components routine and compare them against ground-truth region clusters to produce the VI measurements. The final VI for each threshold is averaged from all the probability maps. For comparison, we also compute segmentations by thresholding the gray value images following the same criteria as the probability maps to produce the VI measurements.  

\begin{figure}[htb]
\begin{minipage}[b]{1.0\linewidth}
  \centering
  \centerline{\includegraphics[clip, trim=0.0cm 0.0cm 0.0cm 0.0cm, width=8.8cm]{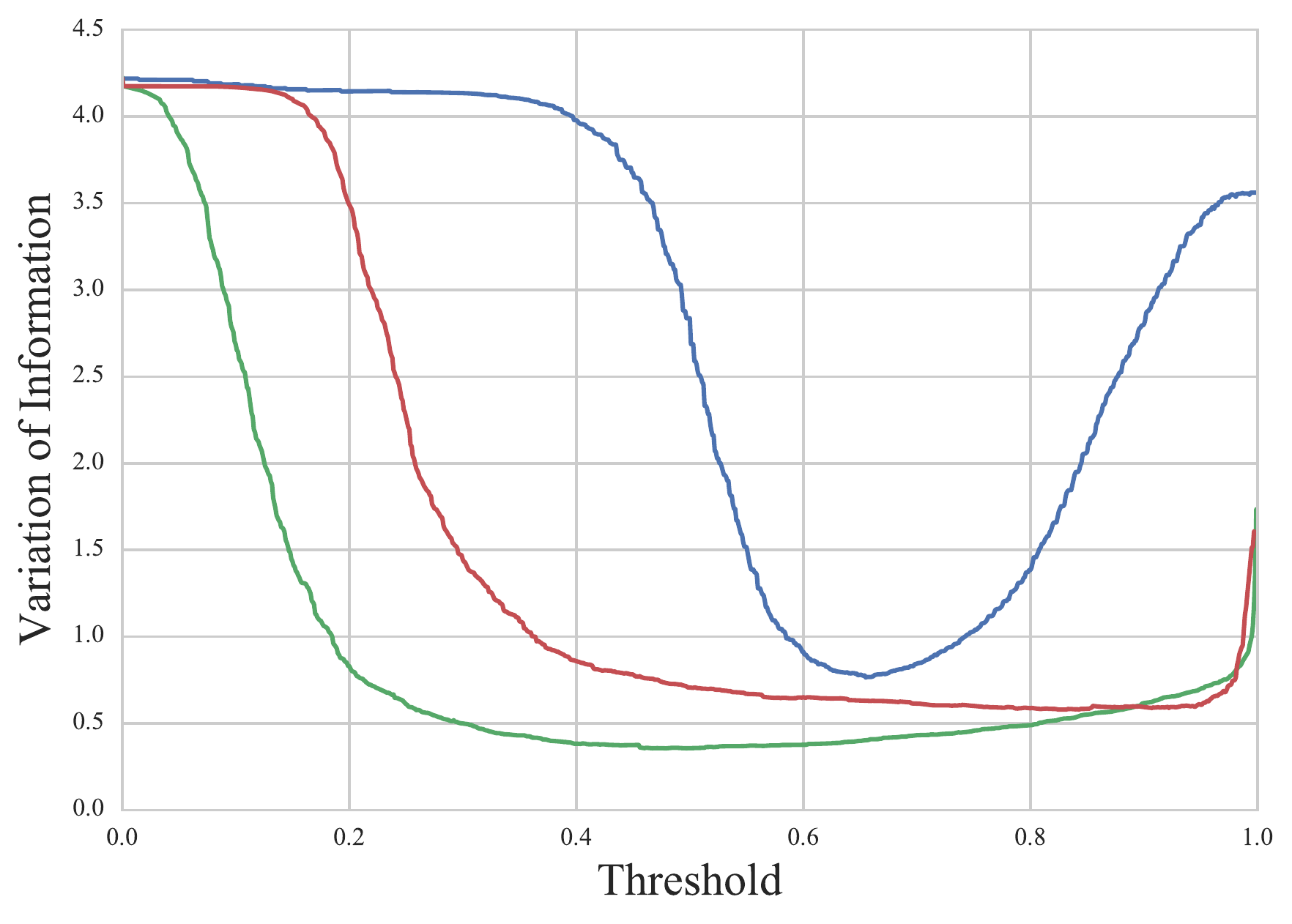}}
\end{minipage}
\caption{Segmentation error in VI on 100 test images. Blue, Red and Green curves show the VI errors for segmentations generated from thresholding gray value images, the output of offline classifier, and the prediction from interactively-trained classifier, respectively.}
\label{fig:vires2}
\end{figure}

Following Kaynig et al.~\cite{Kaynig2015} and Nunez-Iglesias et al. \cite{Nunez-Iglesias13}, we plot the VI curves of segmentations produced by thresholding at different values of CNN pixel prediction in Fig. \ref{fig:vires2}.
These are results we generated from 100 test images.  The blue graph is the results of thresholding the gray value images.  The red graph, shown in the middle, represents the offline CNN classifier. The green graph is the interactive classifier, which achieved the lowest VI value of 0.36. Overall, the interactive classifier led to better results than the others consistently over a large range of thresholds.

The results demonstrates the strength of our interactive training framework.  With sparsely annotated pixels, we can label neuronal structures in EM images and achieve better results than the conventional method of manually labeling all pixels.  


\section{Conclusions}
\label{sec:conclusion}
We presented an interactive approach for training a CNN for segmentation of EM images.  We demonstrated that by training a classifier with sparse annotations using our technique we are able to produce better results than a classifier of the same network architecture trained offline on all ground-truth samples.

We primarily developed the system for applications in neuroscience to address the manual and tedious process of labeling neuronal structures in brain images. We believe that the resulting method is applicable to segmentation tasks beyond the field of neuroscience.  We will make the software of this system freely available upon acceptance of this paper to a peer-reviewed publication.

Future work will concentrate on addressing the training start-up of the interactive system.  Currently, the system requires 100,000 samples before meaningful results can be seen from the network.  A possible remedy to this problem is to use a pre-trained offline network as a seed for the interactive system.

\section{Acknowledgement}
\label{sec:acknowledgement}
This work is partially supported by NSF grants IIS-1447344 and IIS-1607800 and the Intelligence Advanced Research Projects Activity (IARPA) via Department of Interior/Interior Business Center (DoI/IBC) contract number D16PC00002.


\bibliographystyle{IEEEbib}
\bibliography{strings,refs}

\end{document}